\documentclass[letterpaper]{article} 
\usepackage[preprint]{aaai2027}  
\usepackage[hyphens]{url}  
\usepackage{graphicx} 
\usepackage{natbib}  
\usepackage{caption} 
\usepackage{multirow}
\usepackage{subcaption}
\usepackage{algorithm}
\usepackage{algorithmic}
\usepackage{amsmath}
\usepackage{pifont}
\usepackage[dvipsnames]{xcolor}

\usepackage{newfloat}
\usepackage{listings}
\DeclareCaptionStyle{ruled}{labelfont=normalfont,labelsep=colon,strut=off} 
\floatstyle{ruled}
\newfloat{listing}{tb}{lst}{}
\floatname{listing}{Listing}
\usepackage{siunitx}
\usepackage{booktabs}
\usepackage{amsfonts}
\usepackage{color}

\newcommand{\xmark}{\textcolor{OrangeRed}{\ding{55}}}
\newcommand{\tmark}{\textcolor{ForestGreen}{\ding{51}}}

\newcommand{\sys}{\textsc{\textbf{CHORD}}}
\title{\sys: Multi-resolution Compression for Multimodal Time Series}
\author{
    Bibhus Luitel\textsuperscript{\rm 1},
    Payal Mohapatra\textsuperscript{\rm 2},
    Subrata Biswas\textsuperscript{\rm 3},
    Ekata Poudel\textsuperscript{\rm 3},\\
    Sunwoong Kim\textsuperscript{\rm 4},
    Qi Zhu\textsuperscript{\rm 4},
    Bashima Islam\textsuperscript{\rm 1}
}
\affiliations{
    \textsuperscript{\rm 1}University of Massachusetts Amherst\\
    \textsuperscript{\rm 2}Arm\\
    \textsuperscript{\rm 3}Worcester Polytechnic Institute\\
    \textsuperscript{\rm 4}Northwestern University
}

\begin{document}

\maketitle

\begin{abstract}
Wearables, mobile devices, and IoT platforms generate multimodal time-series data continuously and at scale, making efficient storage and transfer a primary concern. The same recordings are reused after collection by many downstream models, often at different temporal resolutions, so a compressed representation should be small, useful across resolutions, and produced at a bit rate known in advance. Prior work meets parts of this: symbolic and error-bounded compressors reconstruct accurately but fix resolution at compression time, learned codecs target one signal domain, and discrete tokenizers reconstruct without accounting for rate. We introduce \sys, a coarse-to-fine compressor for heterogeneous multimodal time series. A residual cascade orders each payload from broad to fine temporal scales and synthesizes corrections through learned shapelet-inspired atoms. A shared pathway supplies a reusable synthesis basis across modalities, while private pathways retain capacity for sensor-specific morphology. Finite scalar quantization fixes the bit width of every modality and scale, so payload size is known before encoding, and scale dropout trains each coarse-to-fine prefix to be a valid reconstruction operating point. Across seven wearable-sensing and forecasting benchmarks, \sys \space stays within 0.62 points of uncompressed classification accuracy on average at the 8$\times$ target, the smallest gap of any compressed method, with all resolutions decoded from one payload.
\end{abstract}

\section{Introduction}
Sensors in wearable, mobile, clinical, and Internet of Things (IoT) systems continuously generate multimodal time-series data. We use modality to denote one sensor stream, such as an accelerometer recording or electrocardiogram channels. Modalities differ substantially in sampling rate, number of channels, amplitude range, frequency content, and waveform morphology. We consider regularly sampled modality windows, including streams buffered and resampled on device. At deployment scale the resulting volume quickly exceeds what can be stored or transferred at low cost: for the WESAD dataset \cite{10.1145/3242969.3242985}, storing day-long chest and wrist recordings would require approximately 2.0\,GB, as the dataset records 5,768 float32 scalars per second. The same recordings are later reused by classifiers, forecasters, and pretrained models with different architectures, motivating compact storage that preserves information across downstream analyses. 

Different analyses require different amounts of temporal detail. Activity recognition sensors are often sampled faster than classification performance requires~\cite{KHAN201633}, and models can be trained to tolerate changes in sampling rate~\cite{Hasegawa_2021}. Adaptive systems exploit this by first processing a coarse signal and invoking a finer-resolution model only when needed~\cite{10271563}. These approaches select resolution during acquisition or within a task-specific model, but do not address how a recording should be stored for later reuse.
Existing storage choices remain limited. Reducing temporal detail before
storage permanently discards information. Storing only a high-fidelity encoding requires every analysis to read the
full representation, while separate encodings per fidelity level
duplicate storage and encoding work. A preferable representation would support progressive refinement from one
stored record.

Existing compression methods provide only parts of this functionality. Signal-reduction methods \cite{645415.652239, 645496.657889, Keogh2001DimensionalityRF, 1795494}, symbolic representations \cite{lin2003symbolic,malinowski20131d,elsworth2020abba}, and error-bounded compressors~\cite{liang2022sz3,chandak2020lfzip} produce compact representations, but generally commit to one configuration and primarily control sample-level approximation. Learned codecs \cite{lee2023vector,zeghidour2021soundstream,defossez2022high} reconstruct
signals through quantized latents, while multimodal representation learning \cite{guo2019deep} produces task-oriented features rather than rate-accounted reconstructive storage. Progressive compressors \cite{10.1145/3588953,Lin2024-cj} provide nested reconstruction levels, but use deterministic constant or linear primitives and do not evaluate whether partial reconstructions preserve downstream neural-model utility. Reusing recordings across models and fidelity levels therefore calls for one compressed record, or payload, whose bit size is known before encoding and whose prefixes give progressively refined reconstructions. The payload aggregates modality-specific codes under one recording-level budget, with fixed stored footprint reported separately.

We introduce \sys, a coarse-to-fine neural compressor for heterogeneous
multimodal time series. First, \sys \space organizes each modality into a residual scale ladder, where
later scales encode detail earlier ones missed. Scale-wise corrections are synthesized from learned shapelet-inspired temporal atoms, and scale dropout trains truncated prefixes as valid reconstruction operating points. Second, shared temporal atoms provide reusable synthesis structure across modalities, while private pathways retain sensor-specific capacity. Third, finite scalar quantization assigns fixed scalar levels to each coefficient dimension, making the payload bit width
predictable once a checkpoint is trained.

Our contributions are as follows.

\begin{itemize}
\item \textbf{Rate-accounted progressive storage.} We formulate time-series compression as a payload measured in bits whose successive prefixes provide multiple reconstruction-fidelity levels without re-encoding.
\item \textbf{Learned coarse-to-fine synthesis.} Unlike progressive compressors based on constant or linear pieces, \sys \space synthesizes residual corrections from learned shapelet-inspired temporal atoms and explicitly trains every prefix to be usable.
\item \textbf{Shared and private multimodal structure.} Existing codecs and tokenizers generally encode each series or modality independently. \sys \space aggregates modality-specific codes into one rate-accounted payload while sharing temporal atoms across modalities and retaining private capacity for sensor-specific morphology.
\end{itemize}

We evaluate \sys \space on seven datasets spanning wearable sensing, activity recognition, and forecasting against signal-reduction, symbolic, error-bounded, progressive, and learned-codec baselines. All methods are evaluated under dataset-level nominal payload budgets, with each modality retaining its native channel count and sampling rate. At the $8\times$ target, \sys \space stays within $0.62$ points of uncompressed
classification accuracy on average, the smallest gap of any method
evaluated. \sys \space also achieves the highest mean compressed accuracy in all sensor-foundation-model dataset--regime comparisons and the lowest CPU latency among learned codecs on all three efficiency datasets, and matches uncompressed forecasting MAE where the learned codecs degrade by three to five times. Each checkpoint supports a finite ladder of prefix operating points from one encoded payload; separate checkpoints are trained for different target compression regimes.

\section{Related Work}

\noindent\textbf{Signal reduction, symbolic, and error-bounded compression.} 
Signal-reduction methods based on piecewise approximation and Fourier truncation provide inexpensive lossy reductions. Symbolic methods \cite{lin2003symbolic,malinowski20131d,elsworth2020abba} convert signals into compact discrete sequences, while SZ3 and LFZip \cite{liang2022sz3,chandak2020lfzip} control reconstruction through user-specified error tolerances. These approaches typically select one operating point before storage and optimize numerical approximation rather
than downstream utility.

\noindent\textbf{Subsequence primitives and dictionary reconstruction.} 
Shapelet methods represent time series using recurring local subsequences,
progressing from search-based discovery to differentiable learned primitives
\cite{rakthanmanon2013fast,grabocka2014learning,liu2025learning}. Their
locality and reuse make them suitable temporal atoms, although conventional
shapelets are primarily discriminative. Convolutional sparse coding
\cite{jas2017learning,dupre2018multivariate} reconstructs signals as
superpositions of shifted atoms, allowing a dictionary to be stored once and
each window represented by its coefficients. Standard formulations, however,
do not discretize these coefficients into rate-controlled progressive
payloads.

\noindent\textbf{Learned codecs.} 
TimeVQ-VAE, SoundStream, and EnCodec \cite{lee2023vector,zeghidour2021soundstream,defossez2022high} combine neural encoders, discrete latents, and learned decoders. SoundStream and EnCodec vary residual-vector-quantization depth over a fixed latent grid rather than organizing refinement by temporal scale. Finite scalar quantization \cite{mentzer2023finite} instead uses fixed per-dimension scalar levels, providing predictable discrete capacity without a learned vector codebook. These codecs are generally designed for individual signal domains rather than heterogeneous modality collections.

\noindent\textbf{Discrete tokenizers and sensor foundation models.} 
TOTEM~\cite{talukder2024totem} trains a convolutional VQ-VAE tokenizer across
time-series domains, while VQShape~\cite{wen2024abstracted} represents series
through abstract temporal shapes and transformation parameters.
Masked-reconstruction sensor models recover masked signal regions or
multiscale coefficients \cite{thukral2026waveletdrivenmaskedmultiscalereconstruction}, and multimodal representation learning~\cite{guo2019deep} handles heterogeneous inputs for prediction. These methods target tokenization or transferable features rather
than progressive storage with explicit payload accounting. Task-oriented communication~\cite{9837474} similarly transmits features optimized for one predetermined inference task.

\noindent\textbf{Progressive compression.} 
HIRE \cite{10.1145/3588953} builds a hierarchical residual encoding serving several accuracy levels from one representation; PBEPLA-RR \cite{Lin2024-cj} stores piecewise-linear residual layers whose partial sums tighten error bounds. Both establish the value of a nested payload, but their hierarchies are deterministic and error-driven, use constant or linear primitives rather than learned synthesis, and are not examined for whether partial payloads preserve what neural models need.

\section{The \sys \space Framework}
\label{section_method}



\begin{figure*}[t]
    \centering
    \includegraphics[width=0.98\textwidth]{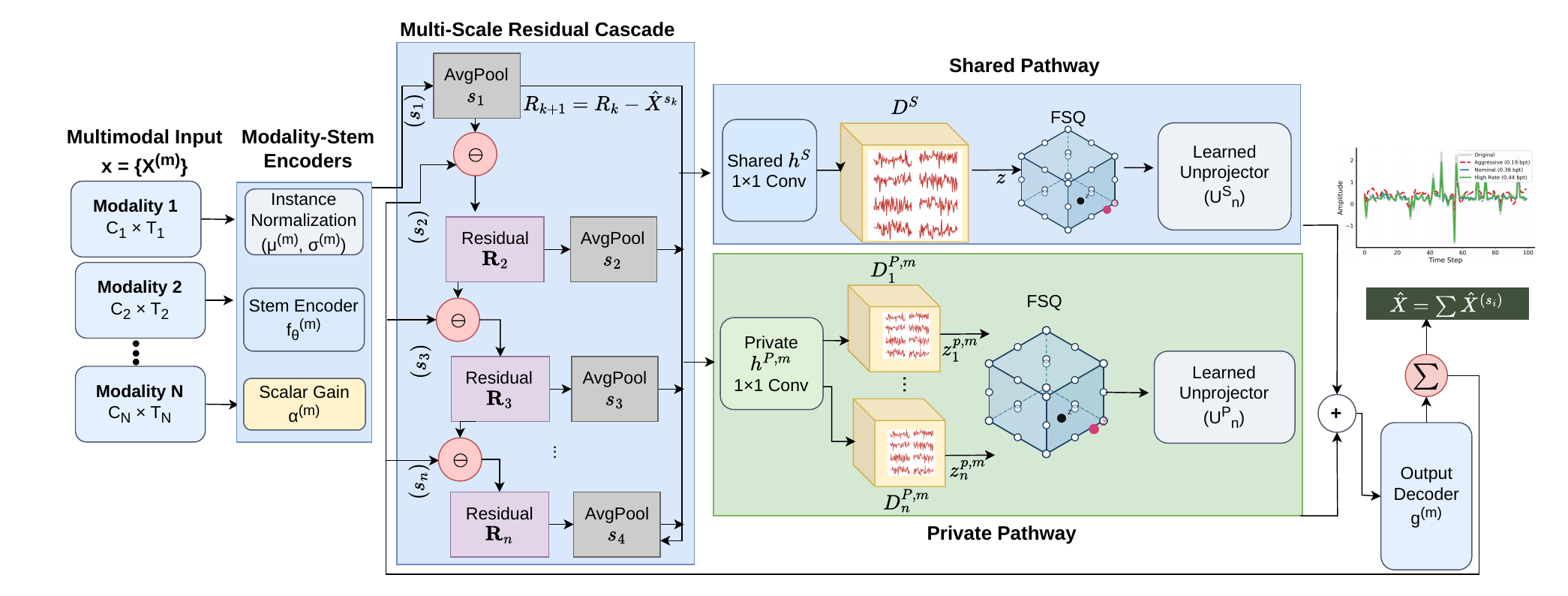}
    \caption{Overview of CHORD. Each modality $X^{(m)}$ is normalized and mapped to a modality-specific latent sequence. At each scale, the pooled residual for each modality is projected into a shared pathway, using the global dictionary $D^S$, and a private pathway, using the modality- and scale-specific dictionary $D_n^{P,m}$. Both coefficient streams are quantized with FSQ, synthesized through their temporal dictionaries, and returned to the stem latent width through learned unprojections. The resulting scale-wise corrections are accumulated and passed through the modality-specific output decoder $g^{(m)}$ to produce $\widehat X^{(m)}$.}
    \label{fig:main}
\end{figure*}

\noindent\sys is a progressive neural compressor for heterogeneous multimodal time series. Let $\mathcal{M}=\{1,\ldots,M\}$ denote a set of $M$ sensing modalities and $\mathbf{x}=\{X^{(m)}\}_{m\in\mathcal{M}}$ a multimodal window, where $X^{(m)}\in\mathbb{R}^{C_m\times T_m}$ has modality-specific channel count $C_m$, sampling rate, and length $T_m$. For each modality, \sys \space produces discrete residual codes over an ordered coarse-to-fine scale ladder $\mathcal{S}^{(m)}=\{s_1,\ldots,s_N\}$, where $s_n\in\mathbb{Z}^{+}$ and with $s_1>\cdots>s_N$, is a latent downsampling factor. At scale $s_n$, the latent sequence of length $T'_m$ is represented using $T_{\mathrm{eff}}^{(m)}(s_n)=\lceil T'_m/s_n\rceil$ code positions, so larger
$s_n$ corresponds to a coarser scale. Decoding an active prefix $\mathcal{A}^{(m)}=\{s_1,\ldots,s_k\}\subseteq\mathcal{S}^{(m)}$ yields an approximation of $X^{(m)}$ that later scales progressively refine. Figure~\ref{fig:main} shows the architecture; implementation details and hyperparameters are provided in Appendix B.

\subsection{Progressive Encoding}
\noindent\textbf{Modality Stems.}
Each modality is normalized independently along time, giving per-window, per-channel statistics $\mu^{(m)},\sigma^{(m)}\in\mathbb{R}^{C_m}$ and
$\widetilde X^{(m)} = (X^{(m)}-\mu^{(m)})/(\sigma^{(m)}+\epsilon)$. A modality-specific stem $f^{(m)}$ maps $\widetilde X^{(m)}$ to $z^{(m)} = \alpha^{(m)}f^{(m)}(\widetilde X^{(m)}) \in\mathbb{R}^{K\times T'_m}$, where $K$ is the common latent width, $T'_m$ is the modality-specific latent length, and the learned scalar $\alpha^{(m)}$ adjusts the latent scale.


\noindent\textbf{Residual Cascade.}
Within each modality, \sys \space encodes $z^{(m)}$ through a coarse-to-fine
residual cascade inspired by Laplacian pyramids \cite{burt1987laplacian}. We initialize
$\bar z^{(m)}_0=0$ and $r^{(m)}_1=z^{(m)}$. At scale $s_n$, the residual is average-pooled to $T_{\mathrm{eff}}^{(m)}(s_n)$ positions, giving $\widehat z^{(m)}_n=\operatorname{AvgPool}(r^{(m)}_n,s_n)$. The pooled residual is then projected, quantized, and synthesized into the scale-specific correction $\operatorname{rec}^{(m)}_n\in\mathbb{R}^{K\times T'_m}$. The cascade updates as $\bar z^{(m)}_n=\bar z^{(m)}_{n-1}+\operatorname{rec}^{(m)}_n$ and $r^{(m)}_{n+1}=z^{(m)}-\bar z^{(m)}_n$.


\noindent\textbf{Shared and Private Pathways.}
At each scale, the pooled residual is mapped through shared and
modality-private projections $W^S_n\in\mathbb{R}^{K^S\times K}$ and
$W^{P,m}_n\in\mathbb{R}^{K^{P,m}\times K}$: \[h^{S,m}_n=W^S_n\widehat z^{(m)}_n, \qquad h^{P,m}_n=W^{P,m}_n\widehat z^{(m)}_n. \] $W^S_n$ is reused across modalities, whereas $W^{P,m}_n$ is modality specific. The codes remain separate per modality, while cross-modal sharing occurs through shared model parameters. Both pathways use finite scalar quantization~\cite{mentzer2023finite}. We write $(\widetilde h^{S,m}_n,c^{(m)}_n)=\operatorname{FSQ}(h^{S,m}_n)$ and $(\widetilde h^{P,m}_n,p^{(m)}_n)=\operatorname{FSQ}(h^{P,m}_n)$, where $\widetilde h$ denotes quantized coefficients and $c,p$ their stored indices. Their implicit vocabulary sizes are $V_n^S=\prod_iL_i^{S,n}$ and $V_n^{P,m}=\prod_iL_i^{P,m,n}$, giving checkpoint-defined bit widths. 

\subsection{Temporal Synthesis and Decoding}
The quantized coefficients are synthesized using a global temporal dictionary
$D^S\in\mathbb{R}^{K^S\times K^S\times L}$, while private codes use a modality-specific dictionary $D^{P,m}\in\mathbb{R}^{K^{P,m}\times K^{P,m}\times L}$, where $L$ is the
atom length. Following convolutional sparse coding \cite{jas2017learning,dupre2018multivariate} and shapelet methods \cite{grabocka2014learning,liu2025learning}, these dictionaries contain learned local temporal atoms optimized for reconstruction. Reusing $D^S$ across modalities encourages temporal atoms that are useful across sensors, while $D^{P,m}$ retains modality-specific capacity. At scale $s_n$, the quantized coefficients are upsampled to $T'_m$ and convolved with their dictionaries, producing shared and private synthesis outputs
$y^{S,m}_n$ and $y^{P,m}_n$. Learned unprojections $U^{S,m}_n\in\mathbb{R}^{K\times K^S}$ and $U^{P,m}_n\in\mathbb{R}^{K\times K^{P,m}}$ return both outputs to the stem
latent width: $\operatorname{rec}^{(m)}_n = U^{S,m}_ny^{S,m}_n+U^{P,m}_ny^{P,m}_n$.
Given an active prefix $\mathcal{A}^{(m)}$, the decoder accumulates these corrections and reconstructs \[\widehat X^{(m)} = \sigma^{(m)} g^{(m)} \left( \sum_{s_n\in\mathcal{A}^{(m)}} \operatorname{rec}^{(m)}_n \right) + \mu^{(m)}. \]
The modality-specific head $g^{(m)}$ is an adapted 1D subpixel decoder \cite{DBLP:conf/cvpr/ShiCHTABRW16} that returns the
latent reconstruction to the native signal grid. Longer prefixes are trained
to improve reconstruction at higher payload cost without re-encoding. One
checkpoint therefore defines a finite ladder of prefix operating points,
while separate checkpoints are trained for each target compression regime. 

\subsection{Payload Rate and Storage Footprint}
At scale $s_n$, the shared and private FSQ codes require $b_n^S=\lceil\log_2V_n^S\rceil$ and $b_n^{P,m}=\lceil\log_2V_n^{P,m}\rceil$ bits per position. The full code payload for a multimodal window is \[ B_{\mathrm{FSQ}}(\mathbf{x}) = \sum_{m=1}^{M} \sum_{s_n\in\mathcal{S}^{(m)}} T_{\mathrm{eff}}^{(m)}(s_n) \left(b_n^S+b_n^{P,m}\right).
\] A prefix payload is obtained by replacing $\mathcal{S}^{(m)}$ with $\mathcal{A}^{(m)}$. Normalization statistics are window dependent and are therefore included in the payload: storing a mean and standard deviation per channel at $b_{\mathrm{norm}}$ bits gives $B_{\mathrm{norm}}(\mathbf{x})=2b_{\mathrm{norm}}\sum_{m=1}^{M}C_m$. The complete payload is $B_{\mathrm{payload}}(\mathbf{x}) =B_{\mathrm{FSQ}}(\mathbf{x})+B_{\mathrm{norm}}(\mathbf{x})$, with compression ratio $\operatorname{CR}_{\mathrm{payload}} =\operatorname{bits}(\mathbf{x})/B_{\mathrm{payload}}(\mathbf{x})$, where $\operatorname{bits}(\mathbf{x})
=b_{\mathrm{raw}}\sum_{m=1}^{M}C_mT_m$. Because $B_{\mathrm{norm}}$ is independent of the code budget, it can dominate at aggressive compression on short windows. Checkpoint parameters and configurations are stored once, reported separately as stored footprint, and excluded from the sample-dependent payload ratio; their cost is amortized across encoded windows.

\subsection{Training}
\noindent\textbf{Scale dropout.} 
At each step and for each modality we sample  prefix length $q_m$ uniformly from $\{1, \ldots, |\mathcal{S}^{(m)}|\}$ and decode only $\mathcal{A}^{(m)} = \{s_1, \ldots, s_{q_m}\}$, so modalities may be truncated at different depths within a step. This exposes the model to truncated payloads during training and prevents it from relying on the finest scale.

\noindent\textbf{Objective.}
Losses are computed in normalized signal space, with $\widetilde{\widehat X}^{(m)} =
(\widehat X^{(m)}-\mu^{(m)})/(\sigma^{(m)}+\epsilon)$: \[ \mathcal{L}_{\mathrm{rec}} =
\sum_{m\in\mathcal{M}} \frac{w_m}{C_mT_m} \left\| \widetilde X^{(m)} - \widetilde{\widehat X}^{(m)} \right\|_2^2 + \lambda_{\mathrm{stft}} \mathcal{L}_{\mathrm{MRSTFT}}, \] where $w_m$ balances modalities and the multi-resolution STFT term is applied to selected modalities~\cite{yamamoto2020parallelwaveganfastwaveform}. After first-stage training, we add Soft-DTW for temporal alignment \cite{cuturi2017soft} and correlation loss for signal shape: $\mathcal{L}_{\mathrm{hybrid}} = \mathcal{L}_{\mathrm{rec}} + \lambda_{\mathrm{dtw}}\mathcal{L}_{\mathrm{SoftDTW}} + \lambda_{\mathrm{corr}}\mathcal{L}_{\mathrm{corr}}$. We regularize quantized coefficient magnitude using $\mathcal{L}_{\mathrm{sparse}} = \sum_m\sum_{n\in\mathcal{A}^{(m)}}(\|\widetilde h^{S,m}_n\|_1+\|\widetilde h^{P,m}_n\|_1)$; this term does not change the fixed-width payload. Scale dropout determines the available prefix, while progressive supervision trains every accumulated latent within that prefix: $\mathcal{L}_{\mathrm{prog}} = \sum_m |\mathcal{A}^{(m)}|^{-1} \sum_{n\in\mathcal{A}^{(m)}} \|\bar z_n^{(m)}-z^{(m)}\|_2^2/(KT'_m)$. Both stages add
$\lambda_1\mathcal{L}_{\mathrm{sparse}} +\lambda_p\mathcal{L}_{\mathrm{prog}}$,
using $\mathcal{L}_{\mathrm{rec}}$ in the first stage and $\mathcal{L}_{\mathrm{hybrid}}$ in the second.

\begin{table*}[t]
\centering
\small
\setlength{\tabcolsep}{0.7mm}

\begin{tabular}{@{}l*{3}{ccc}@{}}
\toprule
&
\multicolumn{3}{c}{\textbf{DaliaHAR}}
&
\multicolumn{3}{c}{\textbf{UCI-HAR}}
&
\multicolumn{3}{c}{\textbf{WESAD}} \\
\cmidrule(lr){2-4}
\cmidrule(lr){5-7}
\cmidrule(lr){8-10}

\textbf{Method}
& \textbf{CR}
& \textbf{Acc.} $\uparrow$
& \textbf{F1} $\uparrow$
& \textbf{CR}
& \textbf{Acc.} $\uparrow$
& \textbf{F1} $\uparrow$
& \textbf{CR}
& \textbf{Acc.} $\uparrow$
& \textbf{F1} $\uparrow$ \\
\midrule

Original
& --
& $84.82_{\pm 1.61}$
& $86.99_{\pm 1.14}$
& --
& $84.18_{\pm 2.70}$
& $85.96_{\pm 2.30}$
& --
& $66.28_{\pm 0.41}$
& $58.41_{\pm 1.89}$ \\
\midrule

Downsampling
& 8.00
& $73.71_{\pm 3.89}$
& $73.58_{\pm 6.04}$
& 8.00
& $30.61_{\pm 0.77}$
& $22.54_{\pm 2.35}$
& 8.00
& $61.98_{\pm 1.68}$
& $55.88_{\pm 2.45}$ \\

Average Pool
& 8.00
& $73.63_{\pm 3.64}$
& $74.13_{\pm 5.44}$
& 8.00
& $44.81_{\pm 4.35}$
& $35.99_{\pm 7.71}$
& 8.00
& $55.46_{\pm 1.39}$
& $46.89_{\pm 2.14}$ \\

PAA
& 8.00
& $72.78_{\pm 2.86}$
& $71.22_{\pm 4.99}$
& 8.00
& $45.24_{\pm 3.96}$
& $36.44_{\pm 6.66}$
& 8.00
& $57.82_{\pm 1.73}$
& $50.75_{\pm 2.04}$ \\

PLA
& 8.00
& $73.74_{\pm 3.07}$
& $74.05_{\pm 4.56}$
& 7.50
& $47.53_{\pm 6.40}$
& $39.89_{\pm 10.40}$
& 8.00
& $57.47_{\pm 0.70}$
& $49.81_{\pm 0.89}$ \\

DFT
& 8.00
& $72.55_{\pm 3.89}$
& $72.10_{\pm 4.81}$
& 7.50
& $47.28_{\pm 5.07}$
& $39.78_{\pm 8.70}$
& 8.00
& $57.18_{\pm 1.72}$
& $48.38_{\pm 2.21}$ \\
\midrule

SAX
& 7.92
& $83.15_{\pm 2.74}$
& $85.15_{\pm 2.37}$
& 7.96
& $\mathbf{83.67_{\pm 0.92}}$
& $\mathbf{85.38_{\pm 0.85}}$
& 7.89
& $62.01_{\pm 4.32}$
& $52.06_{\pm 6.30}$ \\

1D-SAX
& 7.93
& $79.26_{\pm 2.87}$
& $80.82_{\pm 2.17}$
& 6.97
& $82.65_{\pm 0.92}$
& $84.52_{\pm 0.51}$
& 7.90
& $62.70_{\pm 4.63}$
& $52.01_{\pm 7.06}$ \\
\midrule

SZ3
& 11.93
& $82.85_{\pm 0.61}$
& $84.76_{\pm 0.58}$
& 7.95
& $47.28_{\pm 5.57}$
& $48.07_{\pm 9.15}$
& 6.51
& $60.03_{\pm 0.46}$
& $52.36_{\pm 0.36}$ \\

LFZip
& 7.82
& $32.01_{\pm 4.26}$
& $20.25_{\pm 3.89}$
& 7.18
& $22.28_{\pm 1.66}$
& $25.90_{\pm 6.10}$
& 8.73
& $34.32_{\pm 3.83}$
& $30.53_{\pm 0.58}$ \\
\midrule

HIRE
& 8.00
& $73.94_{\pm 1.49}$
& $72.93_{\pm 2.36}$
& 8.00
& $62.16_{\pm 8.60}$
& $58.05_{\pm 9.50}$
& 7.96
& $40.66_{\pm 4.77}$
& $35.60_{\pm 6.02}$ \\
\midrule

TimeVQ-VAE
& 7.98
& $71.97_{\pm 1.47}$
& $76.06_{\pm 1.01}$
& 7.92
& $62.33_{\pm 12.96}$
& $64.97_{\pm 13.83}$
& 7.87
& $57.81_{\pm 2.97}$
& $55.13_{\pm 2.66}$ \\

SoundStream
& 7.99
& $70.25_{\pm 7.71}$
& $71.54_{\pm 4.40}$
& 7.94
& $49.15_{\pm 5.87}$
& $48.51_{\pm 5.90}$
& 7.99
& $51.18_{\pm 0.84}$
& $40.49_{\pm 1.71}$ \\

EnCodec
& 7.96
& $80.21_{\pm 2.58}$
& $82.47_{\pm 1.63}$
& 7.42
& $82.14_{\pm 6.42}$
& $82.79_{\pm 8.00}$
& 7.94
& $59.16_{\pm 1.22}$
& $53.13_{\pm 1.97}$ \\
\midrule

\textbf{\sys}
& 8.29
& $\mathbf{83.62_{\pm 2.18}}$
& $\mathbf{85.83_{\pm 1.84}}$
& 8.10
& $83.59_{\pm 1.64}$
& $84.90_{\pm 1.44}$
& 8.81
& $\mathbf{66.20_{\pm 1.80}}$
& $\mathbf{57.20_{\pm 2.70}}$ \\

\bottomrule
\end{tabular}

\caption{Downstream classification at the nominal $8\times$ compression target. MAESTRO is trained on raw signals and evaluated on reconstructed signals. CR is the realized aggregate compression ratio across all modalities. Values are mean $\pm$ standard deviation over three runs. Best compressed accuracy and macro-F1 per dataset in bold.}
\label{table_multimodal_classification}
\end{table*}
\section{Experiments}
\label{section_experiment}

\subsection{Experimental Setup}

\textbf{Datasets.}
We evaluate \sys \space on seven multivariate time-series datasets spanning wearable physiological sensing (WESAD, DaliaHAR \cite{ppg-dalia_495}), human activity recognition (UCI-HAR \cite{human_activity_recognition_using_smartphones_240}, HHAR \cite{heterogeneity_activity_recognition_344}, SHOAIB\cite{s140610146}, MotionSense\cite{Malekzadeh_2018}), and long-horizon forecasting (ETTh1 \cite{zhou2021informerefficienttransformerlong}). They vary in modality count, channel count, sampling rate, and waveform morphology; per-dataset statistics are in Appendix A. Three downstream task families are evaluated: multimodal classification with MAESTRO \cite{mohapatra2025maestro} on WESAD, DaliaHAR, and UCI-HAR; sensor foundation-model evaluation with LIMU-BERT \cite{xu2025experience} on HHAR, MotionSense, UCI-HAR, and SHOAIB; and forecasting with AutoARIMA and Chronos \cite{ansari2024chronos} on ETTh1. Reconstruction is evaluated on all seven (Appendix C).

\noindent\textbf{Baselines.}
We compare against signal-reduction methods (downsampling, average pooling, PAA~\cite{Keogh2001DimensionalityRF}, PLA~\cite{645496.657889}, and DFT truncation \cite{645415.652239}), symbolic methods (SAX and 1D-SAX), error-bounded compressors (SZ3 and LFZip), the progressive residual compressor HIRE, and neural codecs (TimeVQ-VAE, SoundStream, and EnCodec). Every baseline operates on each modality at its native channel count, sampling rate, and window length, and the reconstructed modalities are reassembled into the dataset's standard multimodal representation. Neural codecs are trained from scratch on the same training split as \textbf{CHORD}, with a separate model per dataset, modality, target ratio, and seed, and no audio-pretrained weights. Each baseline therefore receives dedicated per-modality capacity, whereas \sys \space serves every modality from one jointly trained checkpoint. Per-method hyperparameter search spaces are in Appendix B.

\noindent\textbf{Metrics.}
Compression is reported as the realized payload compression ratio $\mathrm{CR}_{\mathrm{payload}}$, computed at the dataset level as total raw bytes over total payload bytes across all modalities rather than as an average of per-modality ratios. Downstream performance is accuracy and macro-F1 for classification and MAE for forecasting. Stored footprint, reported separately from the payload, is the persistent on-disk footprint in MiB; it's method specific details are found in Appendix B . Efficiency is total encoding-plus-decoding latency per window on CPU and GPU. Appendix C gives rate--distortion results (SNR and NMSE), peak GPU memory, and per-dataset breakdowns.

\noindent\textbf{Evaluation Protocol.}
We evaluate all methods at Low ($8\times$), High ($100\times$), and Aggressive ($500\times$) target ratios. For each dataset and target, we derive a common payload budget from the total raw size of all modalities, and each method searches its rate-controlling parameters for the configuration closest to that budget: signal-reduction and symbolic methods vary retained samples, segments, coefficients, and alphabet sizes; SZ3 and LFZip vary error tolerances; HIRE varies its residual-encoding configuration; and neural codecs vary encoder reduction, latent length, quantizer count, and codebook size. The payload includes any side information needed for decoding, such as segment boundaries, alphabet breakpoints, or normalization statistics; realized ratios for every method, dataset, and regime are in Appendix B. Downstream models are trained on raw signals and tested on reconstructed signals, measuring whether a compressed archive can serve models that were never adapted to it. \sys \space results use the full payload from the checkpoint trained for that target rather than a truncated prefix of a finer checkpoint.

\subsection{Downstream Task Performance}

\textbf{Multimodal Classification.}
Table~\ref{table_multimodal_classification} reports classification at the nominal $8\times$ target. \sys \space is the only method that stays close to uncompressed performance on all three datasets. It gives the best compressed accuracy and macro-F1 on DaliaHAR ($83.62\%$, $85.83\%$) and WESAD ($66.20\%$, $57.20\%$), and is second on UCI-HAR ($83.59\%$, $84.90\%$), where SAX leads by $0.08$ and $0.48$ points, well inside the reported deviations. Averaged over the three datasets, the accuracy gap to uncompressed is $0.62$ points for \sys \space against $2.15$ for SAX, the next closest method. Baselines that approach \sys \space on one dataset fail on another: SAX loses $4.27$ points on WESAD, SZ3 is within $2$ points on DaliaHAR but loses $36.90$ on UCI-HAR, and EnCodec, the strongest neural codec, trails by $7.12$ points on WESAD. \sys \space also reaches these results at or above the nominal target on every dataset ($8.29$, $8.10$, $8.81$), using $1$ to $10\%$ fewer bits than SAX. Error-bounded compressors deviate most from the target because they control error tolerance rather than rate directly, with SZ3 realizing $11.93\times$ on DaliaHAR and $6.51\times$ on WESAD. Consistency across heterogeneous datasets, rather than a win on any single benchmark, is what distinguishes \sys \space here.

\noindent\textbf{Sensor Foundation Model.}
Table~\ref{table_limubert} reports LIMU-BERT accuracy across three compression regimes on four HAR datasets. \sys \space gives the highest compressed accuracy in all twelve dataset--regime comparisons. At the low regime it nearly matches the original signal on UCI-HAR ($88.76\%$ vs.\ $89.00\%$) and stays within $5.3$ points on HHAR and Shoaib, though MotionSense loses $16.6$ points, and it leads the strongest baseline by $3.90$ to $15.54$ points. The advantage widens under stronger compression: margins reach $32.7$ points at the high regime and $44.1$ at the aggressive regime on HHAR, where every baseline falls below $37\%$ while \sys \space holds near $70\%$. Appendix D reports the F1.
\begin{table}[t]
\centering
\setlength{\tabcolsep}{1mm}
\renewcommand{\arraystretch}{1.08}

\begin{tabular}{@{}llcccc@{}}
\toprule
\textbf{ }
& \textbf{Reg.}
& \textbf{HHAR}
& \textbf{MotionSense}
& \textbf{UCI-HAR}
& \textbf{Shoaib} \\
\midrule

O
& --
& 81.00
& 91.00
& 89.00
& 84.00 \\
\midrule

{\shortstack{T}}
& L
& $\underline{71.86}_{\scriptscriptstyle\pm 10.03}$
& $63.04_{\scriptscriptstyle\pm 4.75}$
& $54.67_{\scriptscriptstyle\pm 4.84}$
& $35.03_{\scriptscriptstyle\pm 3.34}$ \\

& H
& $17.31_{\scriptscriptstyle\pm 2.48}$
& $37.76_{\scriptscriptstyle\pm 4.92}$
& $19.42_{\scriptscriptstyle\pm 7.03}$
& $31.90_{\scriptscriptstyle\pm 3.24}$ \\

& A
& $17.11_{\scriptscriptstyle\pm 1.32}$
& $22.67_{\scriptscriptstyle\pm 1.16}$
& $15.62_{\scriptscriptstyle\pm 3.75}$
& $\underline{28.61}_{\scriptscriptstyle\pm 1.57}$ \\
\midrule

{\shortstack{S}}
& L
& $49.21_{\scriptscriptstyle\pm 3.83}$
& $\underline{67.05}_{\scriptscriptstyle\pm 1.92}$
& $66.96_{\scriptscriptstyle\pm 1.27}$
& $\underline{63.85}_{\scriptscriptstyle\pm 2.37}$ \\

& H
& $33.83_{\scriptscriptstyle\pm 2.36}$
& $\underline{60.79}_{\pm 0.82}$
& $\underline{65.81}_{\scriptscriptstyle\pm 1.67}$
& $\underline{56.61}_{\scriptscriptstyle\pm 3.42}$ \\

& A
& $18.00_{\scriptscriptstyle\pm 13.56}$
& $\underline{32.04}_{\scriptscriptstyle\pm 5.63}$
& $37.58_{\scriptscriptstyle\pm 2.44}$
& $28.32_{\scriptscriptstyle\pm 8.89}$ \\
\midrule

{R}
& L
& $36.11_{\scriptscriptstyle\pm 1.18}$
& $43.93_{\scriptscriptstyle\pm 1.90}$
& $\underline{78.93}_{\scriptscriptstyle\pm 2.28}$
& $22.78_{\scriptscriptstyle\pm 0.00}$ \\

& H
& $23.72_{\scriptscriptstyle\pm 1.25}$
& $9.81_{\scriptscriptstyle\pm 1.24}$
& $43.27_{\scriptscriptstyle\pm 3.30}$
& $21.62_{\scriptscriptstyle\pm 1.34}$ \\

& A
& $11.98_{\scriptscriptstyle\pm 0.99}$
& $9.81_{\scriptscriptstyle\pm 1.25}$
& $\underline{62.02}_{\scriptscriptstyle\pm 3.36}$
& $18.00_{\scriptscriptstyle\pm 1.17}$ \\
\midrule

{E}
& L
& $54.24_{\scriptscriptstyle\pm 4.48}$
& $52.54_{\scriptscriptstyle\pm 0.86}$
& $32.40_{\scriptscriptstyle\pm 1.71}$
& $49.83_{\scriptscriptstyle\pm 0.85}$ \\

& H
& $\underline{36.92}_{\scriptscriptstyle\pm 1.86}$
& $39.90_{\scriptscriptstyle\pm 8.66}$
& $29.43_{\scriptscriptstyle\pm 1.57}$
& $47.01_{\scriptscriptstyle\pm 0.38}$ \\

& A
& $\underline{25.61}_{\scriptscriptstyle\pm 0.59}$
& $30.11_{\scriptscriptstyle\pm 1.56}$
& $24.84_{\scriptscriptstyle\pm 3.43}$
& $19.91_{\scriptscriptstyle\pm 0.99}$ \\
\midrule

{\textbf{C}}
& L
& $\mathbf{75.76}_{\scriptscriptstyle\pm 2.09}$
& $\mathbf{74.36}_{\scriptscriptstyle\pm 0.88}$
& $\mathbf{88.76}_{\scriptscriptstyle\pm 0.20}$
& $\mathbf{79.39}_{\scriptscriptstyle\pm 0.18}$ \\

& H
& $\mathbf{69.65}_{\scriptscriptstyle\pm 0.45}$
& $\mathbf{65.72}_{\scriptscriptstyle\pm 0.63}$
& $\mathbf{72.60}_{\scriptscriptstyle\pm 0.92}$
& $\mathbf{74.67}_{\scriptscriptstyle\pm 1.77}$ \\

& A
& $\mathbf{69.72}_{\scriptscriptstyle\pm 0.63}$
& $\mathbf{65.98}_{\scriptscriptstyle\pm 0.33}$
& $\mathbf{72.92}_{\scriptscriptstyle\pm 1.35}$
& $\mathbf{74.46}_{\scriptscriptstyle\pm 1.07}$ \\
\bottomrule
\end{tabular}

\caption{Sensor foundation-model evaluation with LIMU-BERT at low (L, $8\times$), high (H, $100\times$), and aggressive (A, $500\times$) targets, trained on raw and evaluated on reconstructed signals. Mean accuracy (\%) $\pm$ standard deviation over three runs. Best compressed result per column and regime in bold, second-best underlined. O: Original, T: TimeVQ-VAE, S: SoundStream, R: HIRE, E: EnCodec, C: \textbf{CHORD}.}
\label{table_limubert}
\end{table}


\noindent\textbf{Forecasting.}
Table~\ref{table_forecasting} reports MAE on reconstructed ETTh1 signals using AutoARIMA and Chronos. Most non-learned methods remain close to the uncompressed reference: average pooling, PAA, PLA, SAX, 1D-SAX, SZ3, and HIRE all fall within one standard deviation of $2.12$ and $2.06$ MAE, as does \sys \space at $1.88$ and $1.94$. Only DFT truncation and LFZip fall outside among the non-learned methods. The learned codec baselines are the exception, degrading by roughly three to five times, with EnCodec reaching $11.29$ and $10.29$. ETTh1 is a compact, homogeneous forecasting benchmark that exercises none of the heterogeneity \sys \space is built for; the result is that a jointly trained neural compressor need not break on such data, as the other neural codecs do. Stored footprint marks the boundary of the approach: at $0.92$ MiB uncompressed, the dataset is far too small for a $26.09$ MiB checkpoint to amortize.

\begin{table}[t]
\centering
\setlength{\tabcolsep}{1mm}

\begin{tabular}{lccc}
\hline
\textbf{Method} &
  \textbf{\begin{tabular}[c]{@{}c@{}}Footprint \\ MiB $\downarrow$\end{tabular}} &
  \textbf{\begin{tabular}[c]{@{}c@{}}AutoARIMA\\ MAE $\downarrow$\end{tabular}} &
  \textbf{\begin{tabular}[c]{@{}c@{}}Chronos \\ MAE $\downarrow$\end{tabular}} \\ \hline
Original                                                 & 0.92   & $2.12_{\pm 0.83}$          & $2.06_{\pm 0.83}$          \\ \hline
Downsampling & 0.12   & $2.97_{\pm 2.17}$          & $2.48_{\pm 1.12}$          \\
\begin{tabular}[c]{@{}l@{}}Avg.\\ Pool\end{tabular}      & 0.12   & $2.33_{\pm 1.55}$          & $2.36_{\pm 0.87}$          \\
PAA                                                      & 0.12   & $2.23_{\pm1.12}$           & $2.10_{\pm 1.10}$          \\
PLA                                                      & 0.12   & $2.55_{\pm 1.44}$          & $2.60_{\pm 1.38}$          \\
DFT                                                      & 0.12   & $5.54_{\pm 3.50}$          & $2.25_{\pm 1.05}$          \\
SAX                                                      & 0.12   & $2.10_{\pm 0.96}$          & $2.13_{\pm 1.30}$          \\
1D-SAX                                                   & 0.12   & $1.98_{\pm 0.89}$          & $2.05_{\pm 1.09}$          \\
SZ3                                                      & 0.10   & $1.97_{\pm 0.95}$          & $2.17_{\pm 0.80}$          \\
LFZip                                                    & 0.05   & $2.65_{\pm 1.32}$          & $3.14_{\pm 1.47}$          \\ \hline
HIRE                                                     & 0.12   & $2.12_{\pm 0.83}$          & $2.00_{\pm 0.77}$          \\ \hline
TimeVQ-VAE    & 36.72  & $7.82_{\pm 2.64}$          & $7.82_{\pm 2.64}$          \\
SoundStream   & 356.03 & $6.91_{\pm 2.15}$          & $5.97_{\pm 2.20}$          \\
EnCodec                                                  & 41.54  & $11.29_{\pm 7.14}$         & $10.29_{\pm 4.52}$         \\ \hline
\textbf{\sys}                             & 26.09  & $\mathbf{1.88_{\pm 0.70}}$ & $\mathbf{1.94_{\pm 0.96}}$ \\ \hline
\end{tabular}

\caption{Long-horizon forecasting on ETTh1 at the nominal $8\times$ target. AutoARIMA and Chronos are evaluated on reconstructed signals. MAE is mean $\pm$ standard deviation across forecast origins. Best compressed result per column in bold.}
\label{table_forecasting}
\end{table}

\begin{table}[t]
\centering
\setlength{\tabcolsep}{1mm}

\begin{tabular}{@{}llccc@{}}
\toprule
\textbf{Dataset}
& \textbf{Method}
& \textbf{\begin{tabular}[c]{@{}c@{}}Footprint\\MiB $\downarrow$\end{tabular}}
& \textbf{\begin{tabular}[c]{@{}c@{}}GPU total\\ms $\downarrow$\end{tabular}}
& \textbf{\begin{tabular}[c]{@{}c@{}}CPU total\\ms $\downarrow$\end{tabular}} \\
\midrule

Dalia
& HIRE
& 411.77
& --
& $8.33_{\pm 0.08}$ \\

HAR
& TVQ-VAE
& 51.27
& $29.73_{\pm 1.74}$
& $205.70_{\pm 3.14}$ \\

& SStream
& 253.83
& $187.83_{\pm 0.16}$
& $439.40_{\pm 5.27}$ \\

& EnCodec
& 18.23
& $\mathbf{28.92_{\pm 0.04}}$
& $98.96_{\pm 2.33}$ \\

& \sys
& \textbf{16.04}
& $43.56_{\pm 0.06}$
& $\mathbf{82.91_{\pm 2.71}}$ \\
\midrule

UCI
& HIRE
& 0.72
& --
& $2.45_{\pm 0.06}$ \\

HAR
& TVQ-VAE
& \textbf{1.69}
& $16.65_{\pm 3.57}$
& $61.86_{\pm 4.24}$ \\

& SStream
& 101.91
& $18.98_{\pm 0.47}$
& $98.70_{\pm 2.12}$ \\

& EnCodec
& 7.29
& $9.92_{\pm 0.08}$
& $47.19_{\pm 0.40}$ \\

& \sys
& 1.99
& $\mathbf{6.51_{\pm 0.09}}$
& $\mathbf{20.59_{\pm 0.69}}$ \\
\midrule

WESAD
& HIRE
& 496.85
& --
& $4.08_{\pm 0.02}$ \\

& TVQ-VAE
& \textbf{18.48}
& $240.00_{\pm 2.13}$
& $200.64_{\pm 12.24}$ \\

& SStream
& 304.60
& $290.99_{\pm 0.86}$
& $253.82_{\pm 5.70}$ \\

& EnCodec
& 106.55
& $142.01_{\pm 2.62}$
& $290.17_{\pm 3.69}$ \\

& \sys
& 21.88
& $\mathbf{50.55_{\pm 0.58}}$
& $\mathbf{75.59_{\pm 0.11}}$ \\






\bottomrule
\end{tabular}

\caption{Inference efficiency at the $8\times$ target. Footprint MiB is the persistent on-disk footprint; GPU and CPU columns report total encoding-plus-decoding latency per window. Bold denotes the best learned-codec result; HIRE is included as a lightweight CPU reference. TVQ-VAE: TimeVQ-VAE, SStream: SoundStream.}
\label{table_efficiency_summary}
\end{table}

\begin{table}[!htb]
\centering
\setlength{\tabcolsep}{1mm}

\begin{tabular}{cccccccc}
\hline
\multicolumn{4}{c}{\textbf{Enabled Components}}     & \multicolumn{4}{c}{\textbf{Reconstruction MSE} $\downarrow$} \\ \hline
\textbf{S} & \textbf{P} & \textbf{D} & \textbf{Dec} & \textbf{Scale 8}  & \textbf{Scale 4}  & \textbf{Scale 2}  & \textbf{Scale 1}  \\ \hline
\tmark        & \xmark         & \xmark         & \xmark           & 0.929             & 0.957             & 0.963             & 1.002             \\
\xmark         & \tmark        & \xmark         & \xmark           & 0.868             & 0.858             & 0.724             & 0.324             \\
\tmark        & \tmark        & \xmark         & \xmark           & 0.973             & 0.979             & 1.042             & 0.279             \\
\tmark        & \tmark        & \tmark        & \xmark           & 3.874             & 3.978             & 4.272             & 2.444             \\
\xmark         & \tmark        & \tmark        & \tmark          & 0.369             & 0.333             & 0.297             & 0.273             \\
\tmark        & \xmark         & \tmark        & \tmark          & 0.572             & 0.555             & 0.529             & 0.514             \\ \hline
\tmark        & \tmark        & \tmark        & \tmark          & \textbf{0.157}    & \textbf{0.093}    & \textbf{0.038}    & \textbf{0.004}    \\ \hline
\end{tabular}

\caption{Architectural component ablation on WESAD (lower MSE is
better). S: shared pathway, P: private pathway, D: scale dropout,
Dec: subpixel decoder.}
\label{table_ablation_scale_drpt}
\end{table}

\begin{table}[t]
\centering
\setlength{\tabcolsep}{1mm}

\begin{tabular}{@{}lccccc@{}}
\toprule
&
\multicolumn{2}{c}{\textbf{Activation (\%)}}
&
\multicolumn{2}{c}{\textbf{Contribution (\%)}}
& \\
\cmidrule(lr){2-3}
\cmidrule(lr){4-5}

\textbf{Scale}
& \textbf{Shared}
& \textbf{Private}
& \textbf{Shared}
& \textbf{Private}
& \textbf{Cosine} \\
\midrule

8 & 56.4 & 43.6 & 38.1 & 61.9 & 0.195 \\
4 & 57.2 & 42.8 & 38.5 & 61.5 & 0.188 \\
2 & 53.4 & 46.6 & 38.8 & 61.2 & 0.187 \\
1 & 52.3 & 47.7 & 39.2 & 60.8 & 0.178 \\

\bottomrule
\end{tabular}

\caption{Shared and private pathway behavior across temporal scales. Activation is normalized latent magnitude, contribution is relative decoded energy, and cosine is the similarity between the two representations.}
\label{table_pathway_analysis}
\end{table}

\begin{table}[t]
\centering
\setlength{\tabcolsep}{0.8mm}

\begin{tabular}{@{}lccccc@{}}
\toprule
\textbf{Modality}
& \textbf{Sh. act.}
& \textbf{Pr. act.}
& \textbf{Sh. ctr.}
& \textbf{Pr. ctr.}
& \textbf{Cosine} \\
\midrule

C ACC
& 34.7\%
& 65.3\%
& 58.4\%
& 41.6\%
& $-0.092_{\pm 0.042}$ \\

C ECG
& 33.6\%
& 66.4\%
& 21.4\%
& 78.6\%
& $0.523_{\pm 0.030}$ \\

C EDA
& 66.0\%
& 34.0\%
& 50.0\%
& 50.0\%
& $0.082_{\pm 0.043}$ \\

C EMG
& 50.6\%
& 49.4\%
& 44.0\%
& 56.0\%
& $-0.148_{\pm 0.011}$ \\

C RESP
& 39.0\%
& 61.0\%
& 34.3\%
& 65.7\%
& $0.463_{\pm 0.139}$ \\

C TEMP
& 87.4\%
& 12.6\%
& 47.7\%
& 52.3\%
& $0.130_{\pm 0.008}$ \\

W ACC
& 35.0\%
& 65.0\%
& 48.7\%
& 51.3\%
& $0.048_{\pm 0.068}$ \\

W BVP
& 26.2\%
& 73.8\%
& 38.5\%
& 61.5\%
& $0.879_{\pm 0.036}$ \\

W EDA
& 71.4\%
& 28.6\%
& 43.6\%
& 56.4\%
& $-0.098_{\pm 0.134}$ \\

W TEMP
& 79.1\%
& 20.9\%
& 5.1\%
& 94.9\%
& $-0.010_{\pm 0.099}$ \\

\bottomrule
\end{tabular}

\caption{Per-modality pathway behavior on WESAD at the finest scale, with columns as in Table~\ref{table_pathway_analysis}. Cosine similarity is mean $\pm$ standard deviation. C: Chest, W: Wrist.}
\label{table_modality_pathways}
\end{table}

\subsection{Reconstruction and Efficiency}
\label{sec:fidelity_resolution_efficiency}

\textbf{Reconstruction Fidelity.}
At the aggressive regime \sys \space reaches the lowest NMSE of any method on DaliaHAR ($0.050$ at $0.068$ bits per sample (BpS)), WESAD ($0.017$ at $0.063$ BpS), and UCI-HAR ($0.040$ at $0.107$ BpS), in each case at the smallest bit rate shown, and is competitive on Shoaib, HHAR, and MotionSense. The margins are largest on the multimodal datasets whose modalities differ most in sampling rate and morphology, consistent with the shared--private allocation mattering most when a limited budget must be divided across heterogeneous signals. ETTh1 is the exception: \sys \space has the highest NMSE there while achieving the lowest forecasting MAE (Table~\ref{table_forecasting}). Reconstruction on this benchmark discards high-frequency variation that neither forecaster uses, so NMSE and downstream utility diverge, which is why we evaluate compressed signals on downstream models rather than on reconstruction error alone. Full rate--distortion curves for SNR and NMSE on all seven datasets are in Appendix C.

\noindent\textbf{Multi-Resolution Decoding.}
\label{sec:multiresolution}
We decode one trained checkpoint at prefix lengths corresponding to scales $1$ through $64$ and report SNR at each. On DaliaHAR, \sys \space holds $27.30$ dB at the coarsest scale and $40.67$ dB at the finest, and ranges from WESAD $28.49$ to $38.99$ dB, while the learned codecs, which must be re-encoded for each rate, stay below $8.1$ dB throughout. HIRE also serves several accuracy levels from one stored representation and is competitive on SNR at $27.37$ to $43.83$ dB on DaliaHAR, but its truncated payloads do not transfer to downstream models: at the low regime it reaches $36.11\%$ and $22.78\%$ LIMU-BERT accuracy on HHAR and Shoaib against $75.76\%$ and $79.39\%$ for \sys \space (Table~\ref{table_limubert}), and it stores $411.77$ MiB on DaliaHAR against $16.04$ for \sys \space (Table~\ref{table_efficiency_summary}). Progressive access alone is therefore not sufficient; the prefixes must be trained to remain usable, which is what scale dropout provides. Per-scale SNR curves are in Appendix C.

\noindent\textbf{Efficiency.}
Table~\ref{table_efficiency_summary} reports latency and stored footprint at the $8\times$ target. Among learned codecs \sys \space has the lowest CPU latency on all three datasets and the lowest GPU latency on two; EnCodec is faster on GPU for DaliaHAR, the dataset with the most modalities, where \sys \space runs a cascade per modality within one checkpoint. HIRE decodes faster on CPU, but stores $411.77$ and $496.85$ MiB on DaliaHAR and WESAD against $16.04$ and $21.81$ for \sys, and loses $12.90$ and $21.60$ macro-F1 points (Table~\ref{table_multimodal_classification}). Peak GPU memory is in Appendix C.

\subsection{Ablation Studies}
\label{sec:ablations}
\textbf{Architectural Components.}
Table~\ref{table_ablation_scale_drpt} ablates the shared and private pathways, scale dropout, and the subpixel decoder on WESAD. Private-only outperforms shared-only at every scale ($0.273$ vs.\ $0.514$ at scale 1), so modality-specific capacity carries most of the reconstruction, and the shared pathway contributes only in combination with the decoder. The full model reaches the lowest MSE at every scale, from $0.157$ at scale 8 to $0.004$ at scale 1. Scale dropout and the subpixel decoder are confounded here: the one configuration with dropout and no decoder is the worst in the table, and we lack a decoder-without-dropout run to separate them. The table establishes that the full combination is necessary, not that each component contributes on its own.

\noindent\textbf{Shared--Private Specialization.}
Table~\ref{table_pathway_analysis} shows the private pathway carrying roughly $61\%$ of decoded energy at every scale despite lower latent activation, so activation magnitude is a poor proxy for contribution. The two diverge most for wrist temperature in Table~\ref{table_modality_pathways}, where $79.1\%$ shared activation yields $5.1\%$ of decoded energy, indicating the modality-specific unprojection suppresses a pathway the encoder still drives. Mean shared--private cosine similarity is low, but per-modality values at the finest scale span $-0.148$ to $0.879$, so the average partly reflects cancellation rather than uniform decorrelation. Specialization is otherwise modality-dependent, with private contributions dominating the physiological channels and the shared pathway leading only for chest accelerometry. The pathways therefore learn distinct representations for most modalities rather than collapsing.

\noindent\textbf{Quantizer Representation.} 
We compare FSQ against EMA-VQ at a matched capacity of 512 codes on WESAD. FSQ activates 376 codes against 70 for EMA-VQ and reaches lower reconstruction MSE ($0.0643$ vs.\ $0.0783$). FSQ reaches that utilization early in training while EMA-VQ stays concentrated on a small set of codes (Appendix C). The choice is nonetheless driven by rate rather than fidelity: fixed scalar levels fix the bit width per position once a checkpoint is trained, so $B_{\mathrm{payload}}$ is computable before encoding. The utilization result shows this guarantee costs nothing in reconstruction quality.


\section{Conclusion}
We presented \sys, a progressive neural compressor for heterogeneous multimodal time series combining a learned residual scale ladder, shared and modality-private temporal synthesis, and fixed-width finite scalar quantization. Treating the compressed record as a payload with a bit budget known before encoding lets one checkpoint serve several reconstruction levels without re-encoding. Under matched budgets \sys \space stays within $0.62$ points of uncompressed classification accuracy on average, the smallest gap among the methods evaluated. Limitations remain: the framework assumes regular sampling and a fixed modality schema, and we do not test missing modalities or cross-device transfer; progressive decoding operates within one checkpoint; the pathway analysis does not guarantee that shared atoms are reused across sensors; and CPU and GPU measurements do not establish wearable deployment. Future work will address irregular streams, entropy-coded payloads, edge evaluation, and policies that select prefix depth from downstream requirements.

\bibliography{aaai2027}

@incollection{burt1987laplacian,
  title={The Laplacian pyramid as a compact image code},
  author={Burt, Peter J and Adelson, Edward H},
  booktitle={Readings in computer vision},
  pages={671--679},
  year={1987},
  publisher={Elsevier}
}

@inproceedings{cuturi2017soft,
  title={Soft-dtw: a differentiable loss function for time-series},
  author={Cuturi, Marco and Blondel, Mathieu},
  booktitle={International conference on machine learning},
  pages={894--903},
  year={2017},
  organization={PMLR}
}

@article{dupre2018multivariate,
  title={Multivariate convolutional sparse coding for electromagnetic brain signals},
  author={Dupr{\'e} la Tour, Tom and Moreau, Thomas and Jas, Mainak and Gramfort, Alexandre},
  journal={Advances in Neural Information Processing Systems},
  volume={31},
  year={2018}
}

@article{elsworth2020abba,
  title={Abba: adaptive brownian bridge-based symbolic aggregation of time series},
  author={Elsworth, Steven and G{\"u}ttel, Stefan},
  journal={Data Mining and Knowledge Discovery},
  volume={34},
  number={4},
  pages={1175--1200},
  year={2020},
  publisher={Springer}
}

@article{ansari2024chronos,
  title={Chronos: Learning the language of time series},
  author={Ansari, Abdul Fatir and Stella, Lorenzo and Turkmen, Caner and Zhang, Xiyuan and Mercado, Pedro and Shen, Huibin and Shchur, Oleksandr and Rangapuram, Syama Sundar and Arango, Sebastian Pineda and Kapoor, Shubham and others},
  journal={arXiv preprint arXiv:2403.07815},
  year={2024}
}

@inproceedings{grabocka2014learning,
  title={Learning time-series shapelets},
  author={Grabocka, Josif and Schilling, Nicolas and Wistuba, Martin and Schmidt-Thieme, Lars},
  booktitle={Proceedings of the 20th ACM SIGKDD international conference on Knowledge discovery and data mining},
  pages={392--401},
  year={2014}
}

@article{lee2023vector,
  title={Vector quantized time series generation with a bidirectional prior model},
  author={Lee, Daesoo and Malacarne, Sara and Aune, Erlend},
  journal={arXiv preprint arXiv:2303.04743},
  year={2023}
}

@inproceedings{lin2003symbolic,
  title={A symbolic representation of time series, with implications for streaming algorithms},
  author={Lin, Jessica and Keogh, Eamonn and Lonardi, Stefano and Chiu, Bill},
  booktitle={Proceedings of the 8th ACM SIGMOD workshop on Research issues in data mining and knowledge discovery},
  pages={2--11},
  year={2003}
}

@article{liu2025learning,
  title={Learning soft sparse shapes for efficient time-series classification},
  author={Liu, Zhen and Luo, Yicheng and Li, Boyuan and Eldele, Emadeldeen and Wu, Min and Ma, Qianli},
  journal={arXiv preprint arXiv:2505.06892},
  year={2025}
}

@inproceedings{rakthanmanon2013fast,
  title={Fast shapelets: A scalable algorithm for discovering time series shapelets},
  author={Rakthanmanon, Thanawin and Keogh, Eamonn},
  booktitle={proceedings of the 2013 SIAM International Conference on Data Mining},
  pages={668--676},
  year={2013},
  organization={SIAM}
}

@inproceedings{chandak2020lfzip,
  title={LFZip: Lossy compression of multivariate floating-point time series data via improved prediction},
  author={Chandak, Shubham and Tatwawadi, Kedar and Wen, Chengtao and Wang, Lingyun and Ojea, Juan Aparicio and Weissman, Tsachy},
  booktitle={2020 Data Compression Conference (DCC)},
  pages={342--351},
  year={2020},
  organization={IEEE}
}

@article{mentzer2023finite,
  title={Finite scalar quantization: Vq-vae made simple},
  author={Mentzer, Fabian and Minnen, David and Agustsson, Eirikur and Tschannen, Michael},
  journal={arXiv preprint arXiv:2309.15505},
  year={2023}
}

@article{liang2022sz3,
  title={Sz3: A modular framework for composing prediction-based error-bounded lossy compressors},
  author={Liang, Xin and Zhao, Kai and Di, Sheng and Li, Sihuan and Underwood, Robert and Gok, Ali M and Tian, Jiannan and Deng, Junjing and Calhoun, Jon C and Tao, Dingwen and others},
  journal={IEEE Transactions on Big Data},
  volume={9},
  number={2},
  pages={485--498},
  year={2022},
  publisher={IEEE}
}

@article{zeghidour2021soundstream,
  title={Soundstream: An end-to-end neural audio codec},
  author={Zeghidour, Neil and Luebs, Alejandro and Omran, Ahmed and Skoglund, Jan and Tagliasacchi, Marco},
  journal={IEEE/ACM Transactions on Audio, Speech, and Language Processing},
  volume={30},
  pages={495--507},
  year={2021},
  publisher={IEEE}
}

@article{guo2019deep,
  title={Deep multimodal representation learning: A survey},
  author={Guo, Wenzhong and Wang, Jianwen and Wang, Shiping},
  journal={Ieee Access},
  volume={7},
  pages={63373--63394},
  year={2019},
  publisher={IEEE}
}

@inproceedings{malinowski20131d,
  title={1d-sax: A novel symbolic representation for time series},
  author={Malinowski, Simon and Guyet, Thomas and Quiniou, Ren{\'e} and Tavenard, Romain},
  booktitle={International Symposium on Intelligent Data Analysis},
  pages={273--284},
  year={2013},
  organization={Springer}
}

@article{jas2017learning,
  title={Learning the morphology of brain signals using alpha-stable convolutional sparse coding},
  author={Jas, Mainak and Dupr{\'e} la Tour, Tom and Simsekli, Umut and Gramfort, Alexandre},
  journal={Advances in Neural Information Processing Systems},
  volume={30},
  year={2017}
}

@article{defossez2022high,
  title={High fidelity neural audio compression},
  author={D{\'e}fossez, Alexandre and Copet, Jade and Synnaeve, Gabriel and Adi, Yossi},
  journal={arXiv preprint arXiv:2210.13438},
  year={2022}
}

@article{mohapatra2025maestro,
  title={MAESTRO: Adaptive Sparse Attention and Robust Learning for Multimodal Dynamic Time Series},
  author={Mohapatra, Payal and Sui, Yueyuan and Pandey, Akash and Xia, Stephen and Zhu, Qi},
  journal={arXiv preprint arXiv:2509.25278},
  year={2025}
}

@inproceedings{xu2025experience,
  title={Experience Paper: Adopting Activity Recognition in On-demand Food Delivery Business},
  author={Xu, Huatao and Zhang, Yan and Gao, Wei and Shen, Guobin and Li, Mo},
  booktitle={Proceedings of the 31st Annual International Conference on Mobile Computing and Networking},
  pages={1015--1028},
  year={2025}
}

@article{KHAN201633,
title = {Optimising sampling rates for accelerometer-based human activity recognition},
journal = {Pattern Recognition Letters},
volume = {73},
pages = {33-40},
year = {2016},
issn = {0167-8655},
doi = {https://doi.org/10.1016/j.patrec.2016.01.001},
url = {https://www.sciencedirect.com/science/article/pii/S0167865516000040},
author = {Aftab Khan and Nils Hammerla and Sebastian Mellor and Thomas Plötz}
}

@article{Hasegawa_2021,
   title={Smartphone Sensor-Based Human Activity Recognition Robust to Different Sampling Rates},
   volume={21},
   ISSN={2379-9153},
   url={http://dx.doi.org/10.1109/JSEN.2020.3038281},
   DOI={10.1109/jsen.2020.3038281},
   number={5},
   journal={IEEE Sensors Journal},
   publisher={Institute of Electrical and Electronics Engineers (IEEE)},
   author={Hasegawa, Tatsuhito},
   year={2021},
   month=Mar, pages={6930–6941} }

@ARTICLE{10271563,
  author={Yang, Guangyu and Zhang, Lei and Bu, Can and Wang, Shuaishuai and Wu, Hao and Song, Aiguo},
  journal={IEEE Journal of Biomedical and Health Informatics}, 
  title={FreqSense: Adaptive Sampling Rates for Sensor-Based Human Activity Recognition Under Tunable Computational Budgets}, 
  year={2023},
  volume={27},
  number={12},
  pages={5791-5802},
  doi={10.1109/JBHI.2023.3321639}}

@article{10.1145/3588953,
author = {Barbarioli, Bruno and Mersy, Gabriel and Sintos, Stavros and Krishnan, Sanjay},
title = {Hierarchical Residual Encoding for Multiresolution Time Series Compression},
year = {2023},
issue_date = {May 2023},
publisher = {Association for Computing Machinery},
address = {New York, NY, USA},
volume = {1},
number = {1},
url = {https://doi.org/10.1145/3588953},
doi = {10.1145/3588953},
journal = {Proc. ACM Manag. Data},
month = may,
articleno = {99},
numpages = {26}
}

@ARTICLE{Lin2024-cj,
  title     = "Progressive Bounded Error Piecewise Linear Approximation with
               resolution Reduction for time series data compression",
  author    = "Lin, Jeng-Wei and Liao, Shih-Wei and Tsai, Yu-Hung and Huang,
               Ching-Che",
  journal   = "Sensors (Basel)",
  publisher = "MDPI AG",
  volume    =  25,
  number    =  1,
  pages     = "145",
  month     =  dec,
  year      =  2024,
  copyright = "https://creativecommons.org/licenses/by/4.0/",
  language  = "en"
}

@misc{yamamoto2020parallelwaveganfastwaveform,
      title={Parallel WaveGAN: A fast waveform generation model based on generative adversarial networks with multi-resolution spectrogram}, 
      author={Ryuichi Yamamoto and Eunwoo Song and Jae-Min Kim},
      year={2020},
      eprint={1910.11480},
      archivePrefix={arXiv},
      primaryClass={eess.AS},
      url={https://arxiv.org/abs/1910.11480}, 
}

@inproceedings{645415.652239,
author = {Agrawal, Rakesh and Faloutsos, Christos and Swami, Arun N.},
title = {Efficient Similarity Search In Sequence Databases},
year = {1993},
isbn = {3540573011},
publisher = {Springer-Verlag},
address = {Berlin, Heidelberg},
booktitle = {Proceedings of the 4th International Conference on Foundations of Data Organization and Algorithms},
pages = {69–84},
numpages = {16},
series = {FODO '93}
}

@inproceedings{645496.657889,
author = {Keogh, Eamonn J. and Chu, Selina and Hart, David and Pazzani, Michael J.},
title = {An Online Algorithm for Segmenting Time Series},
year = {2001},
isbn = {0769511198},
publisher = {IEEE Computer Society},
address = {USA},
booktitle = {Proceedings of the 2001 IEEE International Conference on Data Mining},
pages = {289–296},
numpages = {8},
series = {ICDM '01}
}

@article{Keogh2001DimensionalityRF,
  title={Dimensionality Reduction for Fast Similarity Search in Large Time Series Databases},
  author={Eamonn J. Keogh and Kaushik Chakrabarti and Michael J. Pazzani and Sharad Mehrotra},
  journal={Knowledge and Information Systems},
  year={2001},
  volume={3},
  pages={263-286},
  url={https://api.semanticscholar.org/CorpusID:15462747}
}

@book{1795494,
author = {Oppenheim, Alan V. and Schafer, Ronald W.},
title = {Discrete-Time Signal Processing},
year = {2009},
isbn = {0131988425},
publisher = {Prentice Hall Press},
address = {USA},
edition = {3rd}
}

@inproceedings{
talukder2024totem,
title={{TOTEM}: Tokenized Time Series Embeddings For General Time Series Analysis},
author={Sabera J Talukder and Yisong Yue and Georgia Gkioxari},
booktitle={ICLR 2024 Workshop on Learning from Time Series For Health},
year={2024},
url={https://openreview.net/forum?id=jzIdR2TXlK}
}

@inproceedings{
wen2024abstracted,
title={Abstracted Shapes as Tokens - A Generalizable and Interpretable Model for Time-series Classification},
author={Yunshi Wen and Tengfei Ma and Tsui-Wei Weng and Lam M. Nguyen and Anak Agung Julius},
booktitle={The Thirty-eighth Annual Conference on Neural Information Processing Systems},
year={2024},
url={https://openreview.net/forum?id=pwKkNSuuEs}
}

@misc{thukral2026waveletdrivenmaskedmultiscalereconstruction,
      title={Wavelet-Driven Masked Multiscale Reconstruction for PPG Foundation Models}, 
      author={Megha Thukral and Cyrus Tanade and Simon A. Lee and Juhyeon Lee and Hao Zhou and Keum San Chun and Migyeong Gwak and Viswam Nathan and Md Mahbubur Rahman and Li Zhu and Mehrab Bin Morshed and Subramaniam Venkatraman and Sharanya Arcot Desai},
      year={2026},
      eprint={2601.12215},
      archivePrefix={arXiv},
      primaryClass={cs.LG},
      url={https://arxiv.org/abs/2601.12215}, 
}

@ARTICLE{9837474,
  author={Shao, Jiawei and Mao, Yuyi and Zhang, Jun},
  journal={IEEE Transactions on Wireless Communications}, 
  title={Task-Oriented Communication for Multidevice Cooperative Edge Inference}, 
  year={2023},
  volume={22},
  number={1},
  pages={73-87},
  doi={10.1109/TWC.2022.3191118}}

@inproceedings{DBLP:conf/cvpr/ShiCHTABRW16,
  author       = {Wenzhe Shi and
                  Jose Caballero and
                  Ferenc Huszar and
                  Johannes Totz and
                  Andrew P. Aitken and
                  Rob Bishop and
                  Daniel Rueckert and
                  Zehan Wang},
  title        = {Real-Time Single Image and Video Super-Resolution Using an Efficient
                  Sub-Pixel Convolutional Neural Network},
  booktitle    = {2016 {IEEE} Conference on Computer Vision and Pattern Recognition,
                  {CVPR} 2016, Las Vegas, NV, USA, June 27-30, 2016},
  pages        = {1874--1883},
  publisher    = {{IEEE} Computer Society},
  year         = {2016},
  url          = {https://doi.org/10.1109/CVPR.2016.207},
  doi          = {10.1109/CVPR.2016.207},
  bibsource    = {dblp computer science bibliography, https://dblp.org}
}

@inproceedings{10.1145/3242969.3242985,
author = {Schmidt, Philip and Reiss, Attila and Duerichen, Robert and Marberger, Claus and Van Laerhoven, Kristof},
title = {Introducing WESAD, a Multimodal Dataset for Wearable Stress and Affect Detection},
year = {2018},
isbn = {9781450356923},
publisher = {Association for Computing Machinery},
address = {New York, NY, USA},
url = {https://doi.org/10.1145/3242969.3242985},
doi = {10.1145/3242969.3242985},
booktitle = {Proceedings of the 20th ACM International Conference on Multimodal Interaction},
pages = {400–408},
numpages = {9},
location = {Boulder, CO, USA},
series = {ICMI '18}
}

@misc{ppg-dalia_495,
  author       = {Reiss, Attila and Indlekofer, Ina and Schmidt, Philip},
  title        = {{PPG-DaLiA}},
  year         = {2019},
  howpublished = {UCI Machine Learning Repository},
  note         = {{DOI}: https://doi.org/10.24432/C53890}
}

@misc{human_activity_recognition_using_smartphones_240,
  author       = {Reyes-Ortiz, Jorge and Anguita, Davide and Ghio, Alessandro and Oneto, Luca and Parra, Xavier},
  title        = {{Human Activity Recognition Using Smartphones}},
  year         = {2013},
  howpublished = {UCI Machine Learning Repository},
  note         = {{DOI}: https://doi.org/10.24432/C54S4K}
}

@misc{heterogeneity_activity_recognition_344,
  author       = {Blunck, Henrik and Bhattacharya, Sourav and Prentow, Thor and Kjrgaard, Mikkel and Dey, Anind},
  title        = {{Heterogeneity Activity Recognition}},
  year         = {2015},
  howpublished = {UCI Machine Learning Repository},
  note         = {{DOI}: https://doi.org/10.24432/C5689X}
}

@Article{s140610146,
AUTHOR = {Shoaib, Muhammad and Bosch, Stephan and Incel, Ozlem Durmaz and Scholten, Hans and Havinga, Paul J. M.},
TITLE = {Fusion of Smartphone Motion Sensors for Physical Activity Recognition},
JOURNAL = {Sensors},
VOLUME = {14},
YEAR = {2014},
NUMBER = {6},
PAGES = {10146--10176},
URL = {https://www.mdpi.com/1424-8220/14/6/10146},
PubMedID = {24919015},
ISSN = {1424-8220},
DOI = {10.3390/s140610146}
}

@inproceedings{Malekzadeh_2018, series={EuroSys ’18},
   title={Protecting Sensory Data against Sensitive Inferences},
   url={http://dx.doi.org/10.1145/3195258.3195260},
   DOI={10.1145/3195258.3195260},
   booktitle={Proceedings of the 1st Workshop on Privacy by Design in Distributed Systems},
   publisher={ACM},
   author={Malekzadeh, Mohammad and Clegg, Richard G. and Cavallaro, Andrea and Haddadi, Hamed},
   year={2018},
   month=Apr, pages={1–6},
   collection={EuroSys ’18} }

@misc{zhou2021informerefficienttransformerlong,
      title={Informer: Beyond Efficient Transformer for Long Sequence Time-Series Forecasting}, 
      author={Haoyi Zhou and Shanghang Zhang and Jieqi Peng and Shuai Zhang and Jianxin Li and Hui Xiong and Wancai Zhang},
      year={2021},
      eprint={2012.07436},
      archivePrefix={arXiv},
      primaryClass={cs.LG},
      url={https://arxiv.org/abs/2012.07436}, 
}


\end{document}